\documentclass[a4paper,twoside]{article}

\usepackage{epsfig}
\usepackage{subcaption}
\usepackage{calc}
\usepackage{amssymb}
\usepackage{amstext}
\usepackage{amsmath}
\usepackage{amsthm}
\usepackage{multicol}
\usepackage{pslatex}
\usepackage{apalike}
\usepackage{graphics}
\usepackage{tabularx}
\usepackage{microtype}
\usepackage{SCITEPRESS}     

\usepackage{booktabs}

\begin{document}
\title{Temporal Bilinear Encoding Network of Audio-Visual Features at Low Sampling Rates}

\author{\authorname{Feiyan Hu, Eva Mohedano, Noel O'Connor and Kevin McGuinness}
\affiliation{Insight Centre for Data Analytics, Dublin City University, Dublin, Ireland}
\email{\{feiyan.hu, eva.mohedano, noel.oconnor, kevin.mcguinness\}@dcu.ie}
}


\abstract{Current deep learning based video classification architectures are typically trained end-to-end on large volumes of data and require extensive computational resources. This paper aims to exploit audio-visual information in video classification with a 1 frame per second sampling rate. We propose Temporal Bilinear Encoding Networks (TBEN) for encoding both audio and visual long range temporal information using bilinear pooling and demonstrate bilinear pooling is better than average pooling on the temporal dimension for videos with low sampling rate. We also embed the label hierarchy in TBEN to further improve the robustness of the classifier. Experiments on the FGA240 fine-grained classification dataset using TBEN achieve a new state-of-the-art (hit@1=47.95\%). We also exploit the possibility of incorporating TBEN with multiple decoupled modalities like visual semantic and motion features: experiments on UCF101 sampled at 1 FPS achieve close to state-of-the-art accuracy (hit@1=91.03\%) while requiring significantly less computational resources than competing approaches for both training and prediction.}

\onecolumn \maketitle \normalsize \setcounter{footnote}{0} \vfill

\section{Introduction}
\label{sec:introduction}
Video contains much richer information than static images.
It is one of the closest projections of real life and it enables many applications like CCTV video analysis, autonomous driving, affective computing, and sentiment analysis.
One feature of video is that it contains temporal context between frames.
Processing speed is also a key issue in video analysis;
in certain scenarios such as live video streaming, accuracy can be compromised to some extent to reduce computational cost.

Many approaches have been proposed for video classification. Popular examples include the two stream model~\cite{simonyan2014two}, ConvNet + LSTM~\cite{carreira2017quo,varol2018long}, 3D ConvNets~\cite{3dconvolutions}, TSN~\cite{wang2016temporal}, TLE~\cite{diba2017deep}, and compressed video action recognition~\cite{wu2018compressed}. All these approaches sample frames at the original frame rate (25-30 FPS), which can incur substantial computational costs at both training and inference time. 
There are also significant hardware requirements needed to train these approaches: the two stream network requires 4 GPUs, the TSN 4 GPUs, the Conv+LSTM and 3D convnets 32/64 GPUs, and TLE 2 GPUs. 
During training the two stream network requires a random sample of 1 RGB frame and 10 optical flow frames, the ConvNet+LSTM a sample of 25 frames, and 3D convnets use a sliding window of 16 RGB frames. 
During testing the two stream network randomly samples 25 RGB and 250 optical flow frames, the ConvNet+LSTM samples 50 frames, 3D convnets use a sliding window of 240 RGB frames, TSN and TLE random samples 75 RGB and 750 optical flow frames. 
Two stream, TSN, TLE, and the compressed method also include 5 corner crops and horizontal flips to augment data by a factor of 10.

Fewer researchers have studied the impact of computationally constraining the video analysis system. In this work, we use a 1~FPS sampling rate and 1 GPU constraint and attempt to maximize accuracy within this computational budget. The motivation for a 1~FPS sampling rate is that nearby video frames are similar 
and we expect that redundant information can be removed using a low frame rate. We also observe that some activities can be accurately predicted by using a small number of images.
We experiment with an extreme fixed mid-frame strategy when training UCF-101, and outperform most other state-of-the-art that uses only RGB frames with a hit@1=84.19\%. 
Under limited computational resources, we propose TBEN to encode and aggregate a long range temporal representation of audio-visual features. Experiments show that it is possible to effectively exploit audio-visual information using a compact video representation. Other decoupled modalities such as motion and label hierarchy should also be used if possible to improve robustness and to compensate for the loss of frames under a low sampling rate.

\section{Related Work}
\label{sec:relatedwork}

\noindent Video processing architectures can be classified, based on how they handle temporal inter-frame dependencies, into those that make an \textit{independence assumption}, those that make a \textit{dependence assumption}, or those that are mixed.

Under the \textbf{independence assumption}, the rank of frame sequence is discarded. Methods like temporal average/max pooling on visual features or classification predictions~\cite{karpathy2014large,yue2015beyond} fall into this category. More sophisticated aggregation techniques such as Bag of Words, VLAD~\cite{arandjelovic2013all}, NetVLAD~\cite{arandjelovic2016netvlad}, ActionVLAD~\cite{girdhar2017actionvlad}, and Fisher Vectors~\cite{sanchez2013image} have also been shown to be effective. Other approaches explore direct pooling strategies, such as max-pooling different temporal resolutions, to construct a global video representation, as used in Temporal Segment Networks~\cite{wang2016temporal}. \cite{diba2017deep} generate a video representation by aggregating temporal information using max pooling, average pooling, or element-wise multiplication, and then use a spatial bilinear model, to encode the aggregated segment representation. Our approach, however, emphasizes the importance of applying bilinear pooling in the temporal domain.\footnote{Note that several approaches include motion as a independent modality instead of encoding temporal dynamics of RGB frames; we include them here because they use the independence assumption considering only the RGB input.}

Under the \textbf{dependence assumption} dynamic information between frames is exploited. Methods have been proposed to either model the inter-frame dynamics or extract features to represent the dynamics. Recurrent neural networks such as LSTMs were an early attempt to model the dynamics of frame sequences~\cite{yue2015beyond,sun2015temporal}, but these models have yet to show improved results over feed-forward architectures that include motion features extracted from optical flow~\cite{carreira2017quo}. 
CNN models can be extended to include 3D kernels to directly model time variations~\cite{3dconvolutions,tran2017convnet}. 
Dynamic temporal information can also be modelled explicitly using two-stream networks.
These networks take the motion information from an optical flow model as a complementary stream to the RGB information~\cite{simonyan2014two}. 
\cite{carreira2017quo} proposed an hybrid two-stream 3D architecture that re-uses ImageNet pre-trained weights by ``inflating'' the weights into 3D kernels. 
Researchers have also investigated a weak dependence assumption using techniques like dynamic images~\cite{bilen2016dynamic}, which use rank pooling to compute a linear combination of all frames (or within a window) to capture longer range temporal information. 
Methods that use 3D convolutions, optical flow on densely sampled frames or RNN are computationally expensive.
In order to compute dense optical flow faster, several methods are proposed to approximate optical flow using neural network such as works in TVNet~\cite{fan2018end} and \cite{piergiovanni2019representation}. 
Some researchers~\cite{wu2018compressed} also use motion vectors from compressed MPEG video for fast classification.

TBEN is based on the independence assumption and uses compact bilinear pooling(CBP) to capture long range temporal correspondences. Bilinear pooling has previously been used various in vision applications~\cite{lin2015bilinear,gao2016compact,Zhang_2019_CVPR} and found to be especially useful for constructing spatial features capable of differentiating between fine-grained categories like breeds of dogs, cars, or aircraft~\cite{yu2018hierarchical}. There are also several works that applying bilinear pooling in video~\cite{hu2018deep,girdhar2017attentional}, and some researchers use bilinear pooling to aggregate features from different modalities~\cite{liu2018towards}. Our approach applies bilinear pooling in the temporal domain.
Another aspect we consider is what accuracy is achievable with a 1~FPS constraint. 
In densely sampled frames, neighbouring frames exhibit considerable redundancy and many can often be safely discarded without significantly impacting performance. 
Only a few methods in the literature directly study the impact of sampling rates on video classification performance.
For example, Yue et al.~\cite{yue2015beyond} evaluated the effect of different temporal resolutions on a 30-frame model with max-pooled convolutional features and conclude that lower frame rates (6~FPS with 30 frames RGB inputs) give higher performance in UCF-101.
\paragraph{FGA-240}\label{sec:fga240}
The Fine Grained Actions 240 introduced by~\cite{sun2015temporal} targets sports videos and labeled with 85 high-level categories from Sports-1M dataset~\cite{karpathy2014large} and 240 fine-grained categories. 
The dataset is split into $48,381$ training videos and $87,454$ evaluation videos.
From original list of YouTube URLs, it was possible to download $\sim60$\% of the original data.
Keyframe extraction was performed uniformly at 1~FPS. In total, the dataset contains frames of $~9$M for training, $~0.9$M for validation and $~3.6$M for testing. A random baseline, which consists of generating random predictions on the downloaded testing partition, obtained Hit@1=$0.4$ and Hit@5=$1.9$. This performance matches with that reported in the original paper which indicates similar distribution between downloaded and original data.
\paragraph{UCF101} UCF101~\cite{soomro2012ucf101} is one of the most commonly used datasets used to test video activity classification. It contains 13,320 videos from 101 action categories. 
We report average performance over the three test splits unless otherwise stated.

\section{Temporal Bilinear Encoding Network}
\label{sec:tben}
\noindent Most previous research has used bilinear pooling for spatial aggregation; here we propose to use bilinear pooling to encode the temporal dimension. We propose several approaches that aggregate frame-level representations into video-level representations to capture long range changes.

\subsection{Aggregating Temporal Information}
\paragraph{Compact Bilinear Pooling} Second-order pooling methods~\cite{tenenbaum2000separating} have been shown to be effective at encoding local spatial information for fine-grained visual recognition tasks using CNN models~\cite{lin2015bilinear,yu2018hierarchical}. In this work, we explore Compact Bilinear Pooling (CBP)~\cite{gao2016compact} as an efficient approximation of bilinear pooling, to capture local spatial and long range temporal structure of video frames in a compact global video representation. 
Second-order pooling or fully bilinear representations are formulated as: 
\begin{equation}
B(X)= \sum_{s \in S}^{|S|} x_s x_s^T,
\end{equation}
%
where $X = \{ x_i \in \mathbb{R}^c : 1 \leq i \leq |S|\}$ represents a set of local descriptors $S$ of dimension $c$. In this case, $X \in \mathbb{R}^{h\times w\times c}$ represents the activations of a convolutional layer, with $x_{s}$ being one of the local features and $|S| = h \times w$. Since the cost of the fully bilinear model is expensive, it is popular to approximate the bilinear kernel using compact approaches such as Random Maclaurin (RM)~\cite{kar2012random} and Tensor Sketch (TS)~\cite{pham2013fast} as proposed in~\cite{gao2016compact}.
The TS method is explored in \cite{diba2017deep} to aggregate a fixed set of frames per video, whereas we focus on RM projections to aggregate a variable length sequence of frames. The RM can be easily implemented using two linear layers following: 
\begin{equation}
f_{RM}(x_s)= \sigma (W_1 \cdot x_{s} \circ W_2 \cdot x_{s}),
\end{equation}
where $W_1, W_2 \in \mathbb{R}^{c\times d}$ are \textit{fixed} random Rademacher matrices sampled uniformly from $\{+1, -1\}$, $d>c$, and $\circ$ represents the Hadamard product. $\sigma$ is a normalization function, which can be signed square root, sigmoid, or any other type of transfer function; we found that signed square root worked best for TCBP in FGA240 and rescaling by $ d \times h \times w \times 7$ in UCF101 in our experiments (the average video length in UCF101 is approx.~7 seconds).

\begin{figure}[ht]
\begin{center}
   \includegraphics[width=1.0\linewidth]{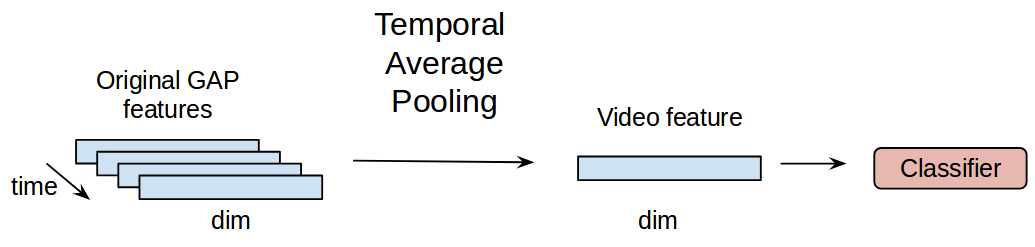}
\end{center}
   \caption{TAP of video frame representations.}
\label{fig:temp_avg}
\end{figure}

\begin{figure}[ht]
\begin{center}
   \includegraphics[width=1.0\linewidth]{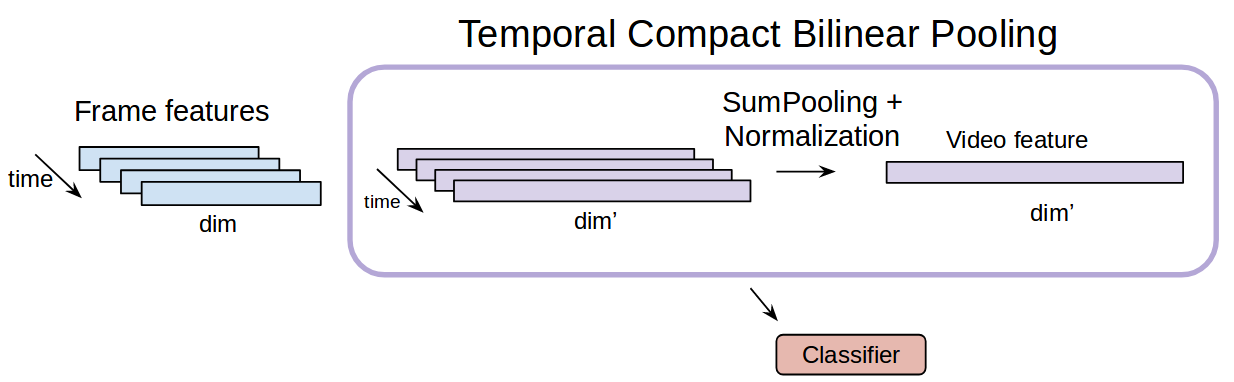}
\end{center}
   \caption{TCBP of video frame representations.}
\label{fig:temp_cbp}
\end{figure}

\paragraph{Temporal and Spatial Information Aggregation} 
Average pooling (\textbf{AP}) is one simple approach to aggregating spatial and/or temporal information from local features in the last convolutional layer. 
The video representation is generated by simply average pooling over time (\textbf{TAP}) and/or space (\textbf{SAP}) over all sampled frames in a video. Figure~\ref{fig:temp_avg} illustrates TAP. 
We propose to use Compact Bilinear Pooling (\textbf{CBP}) to aggregate over the spatial and/or temporal dimension in video. Temporal Compact Bilinear Pooling (\textbf{TCBP}) aggregates information across time: the frame-level representations at each time point are projected to a higher dimensional representation using CBP, then are sum-pooled over the time to generate a global video representation (see Figure~\ref{fig:temp_cbp}).

When processing video clips with a 2D CNN, each frame independently ignoring the temporal dimension to produce a representation $\hat{X} \in \mathbb{R}^{t \times h\times w\times c}$ from last convolutional layer, which is then used as input to TBEN. To transform this into a compact spatial-temporal representation, the information of $\hat{X}$ on spatial ($h$ and $w$) and temporal ($t$) dimensions needs to be aggregated. There are several approaches that could be used to achieve this. Spatial CBP (SCBP) has been shown to be more effective than a fully connected layer to aggregate features from convolution layers in fine-grained visual recognition tasks~\cite{lin2015bilinear}. SCBP using the RM approximation can be defined as:
\begin{equation}
f_s(\hat{X}) = \sum_{s \in H\times W}^{|S|} f_{RM}(\hat{x}_s),
\end{equation}
where $f_s:\mathbb{R}^{t\times h\times w\times c} \rightarrow \mathbb{R}^{t\times c}$. Temporal CBP (TCBP) is defined similarly:
\begin{equation}
f_t(\hat{X}) = \sum_{s \in T}^{|S|} f_{RM}(\hat{x}_s),
\end{equation}
where $f_t:\mathbb{R}^{t\times h\times w\times c} \rightarrow \mathbb{R}^{h\times w\times c}$.
TCBP and SCBP can act as two independent modules. For example, we can use SCBP to pool spatial information and then use TAP or TCBP to aggregate temporal information. We can also discard SCBP, by just using Global Average Pooling of last convolutional layer and then apply TAP or TCBP. Finally, we can also pool spatial-temporal information using CBP jointly, which we refer to as \textbf{STCBP}:
\begin{equation}
\label{f_st}
f_{st}(\hat{X}) =  \sum_{s \in T\times H\times W}^{|S|} f_{RM}(\hat{x_s}),
\end{equation}
where $f_{st}:\mathbb{R}^{t\times h\times w\times c} \rightarrow \mathbb{R}^{c}$. This can give a different representation to applying SCBP and then TCBP (SCBP-TCBP), which is defined as:
\begin{equation}
\label{f_t_f_s}
f_t(f_s(\hat{X})) = \sum_{s \in T}^{|S|} f_{RM}\left(\sum_{s \in H\times W}^{|S|} f_{RM}(\hat{x_s})\right),
\end{equation}
where $f_s:\mathbb{R}^{t\times h\times w\times c} \rightarrow \mathbb{R}^{t\times c}$ and $f_t:\mathbb{R}^{t\times c} \rightarrow \mathbb{R}^{c}$. To make the abbreviation clear, we use the notation $f_{st}(\cdot)$ and $f_t(f_s(\cdot))$ for Equation~\ref{f_st} and \ref{f_t_f_s}. We also define $f_t(f_{\Bar{s}}(\cdot))$, $f_{\Bar{t}}(f_s(\cdot))$ and $f_{\Bar{st}}(\cdot)$ for SAP-TCBP, SCBP-TAP, and STAP. In our experiments where spatial and temporal pooling are applied sequentially, spatial pooling is always applied first.

\subsection{Hierarchical Label Loss}
\label{sec:hierarchy}

\begin{figure}[t]
\begin{center}
   \includegraphics[width=1.0\linewidth]{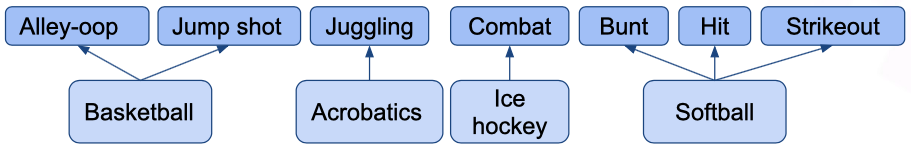}
\end{center}
   \caption{Example illustrating the coarse and fine level annotations in the FGA-240 dataset.}
\label{fig:hier}
\end{figure}

We explore the label dependency by combining the classification of coarse and fine-grained categories, similarly to how it is used in the YOLO object detector network~\cite{redmon2017yolo9000}.
Figure~\ref{fig:hier} shows an example of fine-grained classes and their parents. Parent classes are displayed on the top row and child classes on the bottom. The joint parent-child class distribution is given by:
%
\begin{equation}
P(A_p, A_c) = P(A_c| A_p) P(A_p),
\label{eg:1}
\end{equation}
where $A_p$ is a random variable representing the label of parent class, $A_c$ a random variable representing the label of the child classes, and $P(A_p, A_c)$ is the joint probability of a video being labeled with a particular parent and child class. 
The classifier is easily implemented using a fully connected layer of 325 neurons (240 child and 85 parents), where a softmax normalization is applied on the parent activations to obtain $P(A_p)$ (only one parent class is possible per video)
, and a softmax normalization is applied on the child activations to obtain $P(A_c| A_p)$ (only one child is possible per video).
The final probability score is obtained by multiplying parent and child probabilities as in Eq.~\eqref{eg:1}.

\subsection{Decoupled Modalities}
\paragraph{Short-term motion}
Many researchers~\cite{carreira2017quo,yue2015beyond,varol2018long}
have reported the importance of motion features, specifically the usefulness of optical flow. Although TBEN is designed to capture and aggregate long range temporal information, short term motion is important for distinguishing between certain activities (e.g.~soccer juggling versus soccer penalties). 
To capture the very short-term motion and complement the long-range temporal information captured by TBEN, we compute optical flow sparsely. Section \ref{sec:ucf101} gives further details.
\paragraph{Audio}
\label{sec:audio}
Audio is an important cue in videos, yet few researchers have used it for video classification. 
VGGish model~\cite{hershey2017cnn} is used to extract 128D audio embeddings at every second. 
It takes 0.38s to extract one minute of audio.
Table~\ref{table:audio_tcbp} shows the performance of the audio modality when setting the TCBP output dimensions to 512D and 4096D.
The higher dimensional representations of audio perform better. Audio alone performs significantly worse than the visual modality ($23.49$\% vs $44.87$\% and $44.33$\% Hit@1; see Table~\ref{table:fga_resnet_inception}).
Some of the audio is unrelated to the semantic video content (e.g. musical scores), which explains the poor performance of audio alone.
\begin{table}[ht]
\centering
\begin{tabularx}{\columnwidth}{Xll}
\toprule
Dimension & Hit@1 & Hit@5   \\ \midrule
512 & 19.52 & 43.17   \\
4096 &23.49 & 47.66   \\
\bottomrule
\end{tabularx}
\caption{Results of TCBP on the audio modality.}
\label{table:audio_tcbp}
\end{table}

\section{Experiment Results}
\label{sec:experiments}
\noindent All single frame models for FGA240 and UCF101 are sampled at 1 FPS.
All experiments in the this and the following sections are performed on a single NVIDIA GeForce GTX TITAN X 12GB GPU.

\subsection{FGA240}
FGA240 is much larger than UCF101, and contains longer videos.
To train classification models in a reasonable length of time, we do not fine-tune the CNNs, but instead use output of last convolutional layers of ResNet-50~\cite{he2016deep} and Inception-v3~\cite{szegedy2016rethinking} as features. 

\paragraph{Single Frame Model}
\label{single_frame}
We first conduct experiments on a single frame model. Here, all sampled video frames are used for training a image classifier. During inference, the average predictions of all frames is used as the final prediction. The current state-of-the-art~\cite{sun2015temporal} on FGA240 uses Alexnet, which is slightly dated. 
Our single frame model uses ResNet-50 and Inception-v3 features and a linear layer with softmax activation to generate the final predictions. 
The linear layer is trained using SGD with momentum 0.9 and learning rate 0.01. Results of single frame model are report in Table~\ref{table:fga_resnet_inception}.

\paragraph{Average and Bilinear Pooling}
Table~\ref{resnet_inception_ap_cbp} shows the results of combinations of average and/or bilinear pooling on spatial and temporal dimensions.
The table shows that AP in both the spatial and temporal dimension gives the poorest performance. With ResNet-50 features CBP on the temporal dimension always improves performance over AP. With Inception-v3 features, however, the results are less conclusive and suggest applying CBP in either the spatial or temporal domain, but not both.

With temporal pooling the training set is reduced from $9$M features to approx $40$K features. The aggregation also makes inference much faster. The average length of test videos in FGA240 is $\sim 130$ seconds meaning that during inference, the most computation time is spent on computing the compact bilinear representation, which takes $\sim 2.5$ milliseconds per video for SAP+TCBP; 1~ms/video for SCBP+TAP; 20~ms/video for STCBP; and 26ms/video for SCBP+TCBP. Once the representations are computed, the classification for each video takes under 1~ms.

\begin{table}[ht]
\resizebox{\columnwidth}{!}{%
\begin{tabular}{lcccc}
\toprule
 & \multicolumn{2}{c}{ResNet-50} & \multicolumn{2}{c}{Inception-v3} \\
 & Hit@1 & Hit@5 & Hit@1 & Hit@5 \\ \midrule
$f_{\Bar{st}}(\cdot)$ & 42.75 & 75.85 & 43.24 & 76.49 \\
$f_t(f_{\Bar{s}}(\cdot))$ & 43.40 & 76.37 & \textbf{44.33} & 76.30 \\
$f_{\Bar{t}}(f_s(\cdot))$ & 44.41 & 77.10 & 44.08 & \textbf{76.76} \\
$f_t(f_s(\cdot))$ & \textbf{44.73} & \textbf{77.41} & 44.00 & 75.54 \\ \bottomrule
\end{tabular}%
}
\caption{Combination of average pooling and bilinear pooling on the spatial and/or temporal dimension on FGA240 using ResNet-50 and Inception-v3 features.}
\label{resnet_inception_ap_cbp}
\end{table}

\paragraph{Comparison with the State-of-the-Art}
Table~\ref{table:fga_resnet_inception} shows the best results using TBEN using ResNet-50 features and Inception-V3. For ResNet-50 features the best configuration is to pool both spatial and temporal information with CBP, giving $44.87$\% Hit@1 and taking just $22$s/epoch to train.
As for Inception-v3 features, the best performance (Hit@1 $44.33\%$) using visual features is achieved using SAP and followed by TCBP.
\begin{table*}[ht]
\centering
\begin{tabularx}{\textwidth}{Xcccccc}
\toprule
  & \multicolumn{1}{c}{Dim} & \multicolumn{2}{c}{Hit@1} & \multicolumn{2}{c}{ Hit@5} & \multicolumn{1}{c}{\begin{tabular}[c]{@{}l@{}}Time/  Epoch\end{tabular}} \\ 
 & & R$^+$ & I$^+$ & R$^+$ & I$^+$ &  \\ \midrule
Single Frame (SAP) & 2048 & 40.27 & 42.21 & 72.26 & 73.63 & c.~0.95h \\ \midrule
TBEN$^*$ & 4096 & 44.87 & 44.33 & 77.41 & 76.30 & c.~20s \\
TBEN + Audio & 4608 & 47.42 & 46.67 & 79.59 & 79.14 & c.~20s\\
TBEN + Hierarchy & 4096 & 45.77 & 44.50 & 78.79 & 78.01 & c.~20s\\ 
TBEN + Audio + Hierarchy & 4608 & 47.95 & 47.20 & 80.73 & 80.29 & c.~50s\\ \midrule
LSTM with LAF~\cite{sun2015temporal} & 2048 & \multicolumn{2}{c}{43.40} & \multicolumn{2}{c}{74.90} & c.~3h \\ \bottomrule
\end{tabularx}
\caption{Test performance using ResNet-50 and Inception-V3 features in FGA240. TBEN$^*$ uses $f_{st}(\cdot)$ and $f_t(f_{\Bar{s}}(\cdot))$ for ResNet-50 and Inception-V3 representations respectively. R$^+$ and I$^+$ represent ResNet-50 and Inception-V3 features.}
\label{table:fga_resnet_inception}
\end{table*}
We use concatenation to combine audio and visual features and it provides the best results (Hit@1=$46.6$\% and $47.4$\%) and greatly boosts the performance over the individual modalities (Table~\ref{table:fga_resnet_inception}, \textit{+ Audio}).
For both Hit@1 and Hit@5, the model with the label hierarchy outperforms the one without (Table~\ref{table:fga_resnet_inception}, \textit{+ Hierarchy}).
If we include TBEN, audio, and the label hierarchy, we achieve Hit@1 of $47.95$ using ResNet-50 and $47.20$ using Inception-v3 features, compared with a previous state-of-the-art of $43.40$, while being substantially more computational efficient.

\paragraph{Comparison with BOVW Encoding}
\label{compare_bovw}
We also compare the performance of TBEN with using other sophisticated bag-of-visual-words methods for pooling, such as VLAD and Fisher Vectors, which are used to aggregate temporal information using average pooled spatial features.
We randomly sample 30K videos and for each video sample 100 frames to run $k$-means or fit Gaussian mixture models.
Table~\ref{table:fga comparing with soa} shows the results of using number of cluster $k=64$ and $128$ for VLAD and Fisher vector.
We notice that 64 clusters performs better than 128 for Fisher vectors. This may be due to the poor convergence of the GMM when when using 128 components.
NetVLAD uses $64$ clusters and an output dimension of $4096$. For each video, $300$ frames of spatially average pooled ResNet-50 features are extracted as input for NetVLAD. The batch size is set to 20 during training. The remaining training parameters are set to be the same as Section~\ref{single_frame}. Table~\ref{table:fga comparing with soa} shows that neural network based BOVW methods such as NetVLAD outperform traditional methods. When comparing NetVLAD and TBEN, we achieve inferior Hit@1 and superior Hit@5 by using spatial AP and then temporal CBP, but TBEN is substantially faster and this is achieved without trainable encoding parameters. It is possible, however, to achieve similar Hit@1 performance and superior Hit@5 performance to NetVLAD by using joint spatial-temporal CBP, but in this case the CBP module has to process more features: $f_{st}(\cdot)$ processes $7\times 7$ more features in the CBP module than $f_t(f_{\Bar{s}}(\cdot))$ resulting in longer encoding times.

\begin{table}[ht]
\centering
\scalebox{1.0}{
\begin{tabular}{lrccc}
\toprule
        & $k$ & Hit@1 & Hit@5 & Time (s)  \\ \midrule
VLAD    &  64 & 35.24 & 68.67 & 0.059     \\
VLAD    & 128 & 35.27 & 67.57 & 0.108     \\ \midrule
FV      &  64 & 39.24 & 73.21 & 0.023     \\
FV      & 128 & 31.51 & 64.45 & 0.039     \\ \midrule
NetVLAD &  64 & 44.61 & 74.82 & 0.007     \\
$f_t(f_{\Bar{s}}(\cdot))$ & N/A & 43.40 & 76.37 & 0.002 \\
$f_{st}(\cdot)$ & N/A & 44.87 & 77.36 & 0.022  \\
\bottomrule
\end{tabular}
}
\caption{Results of VLAD, Fisher Vectors, and NetVLAD encoding schemes. Time shown is the average time in seconds to encode the features for a single video. }
\label{table:fga comparing with soa}
\end{table}

\subsection{UCF101}
\label{sec:ucf101}
In the training process, we fine-tune the ResNet-50 in an end-to-end manner with TBEN embedded after feature extraction. 
\paragraph{Sampling Rate}
We propose a new mid-frame sampling strategy, which only takes the middle frame for each 1~FPS sampled video during training. 
During testing, all frames sampled at 1~FPS are processed, computing the average of the individual predictions. 
Inception V3 is used to train the single frame model using SGD with a momentum of 0.9 and base layers of 0.01 on the last linear and auxiliary linear layer and 0.001 elsewhere. 
The results indicate that, even with the significant data reduction, we still achieve quite reasonable accuracy (84.19\% in Table~\ref{tab:ucf101_results}).
\begin{table}[ht]
\centering
\begin{tabular}{lllll}
\toprule
FPS & 0.5 & 1 & 2 & 4 \\ \midrule
SCBP + TCBP & 85.75 & 86.41 & 84.85 &  74.02\\ \bottomrule
\end{tabular}
\caption{Accuracy of SCBP + TCBP using different sampling rates on UCF101.}
\label{resampling_ucf101}
\end{table}
To experiment different frame sampling rate, we use 7s sliding window with stride of 2s in training and 4s in testing.
Table~\ref{resampling_ucf101} shows the results of using TBEN to aggregate temporal features using different sampling rates. 
We see that TBEN does not improve with increased sampling rates.
In fact increasing the sampling rate to 4~FPS results in a performance decrease of about 10\% comparing with 2~FPS.
This suggests that TBEN is good at capturing long range information, but that small variations between nearby frames might cause problems. 

\paragraph{Comparison with the State-of-the-Art}
Following other state-of-the-art, for motion feature we use optical flow to encode short-term motion with Farneback's dense optical flow \cite{farneback2003two} in OpenCV over 5 consecutive frames using 1FPS sampling rate. 
Weights of the backbone network are initialized as in \cite{wang2016temporal}. This modality alone achieves 67.19\% accuracy.
A pretrained VGGish network is used to extract 128D audio features that are the same length as the video with 1 descriptor per second. TCBP is used to aggregate all temporal audio features to a 1024D representation, achieving 24.09\% accuracy. 
Each modality: RGB, TSCBP, motion, and audio are trained independently on one GPU. The final predictions are fused by combining the activations of the last linear layers. Table~\ref{tab:ucf101_results} lists the boost from each modality when we add them in sequential order: RGB, motion, and audio.
Static visual features from the fixed mid-frame model give approximately a 1\% boost on top of RGB. 
Adding optical flow gives another 2.67\% boost, which is the largest among the added modalities.
Audio features are fast to compute and give an almost 2\% boost.
\begin{table}[ht]
\centering
\scalebox{1.0}{
\begin{tabularx}{\columnwidth}{Xllll}
\toprule
 & Split 1 & Split 2 & Split 3 & Mean \\ \midrule
TBEN* & 85.25 & 85.22 & 86.74 & 85.74 \\ 
RGB & 83.35 & 84.09 & 85.12 & 84.19 \\ 
OF & 66.40 & 66.60 & 68.56 & 67.19 \\ 
Audio & 24.13 & 24.58 & 23.57 & 24.09 \\ 
All & 91.44 & 90.63 & 91.02 & 91.03 \\ 
\midrule
+RGB &+1.06&+1.23& +0.76 & +1.02\\ 
+OF &+3.25&+2.46& +2.30 & +2.67\\ 
+Audio &+1.88&+1.72&+2.22& +1.94\\ \bottomrule
\end{tabularx}
}
\caption{Accuracy on UCF101 using different modalities and accuracy gains by adding modalities in late fusion. TBEN* using STCBP with image size $224\times224$.}
\label{tab:ucf101_results}
\end{table}

\begin{table}[ht]
\centering
\resizebox{\columnwidth}{!}{%
\begin{tabular}{lll}
\toprule
 & \begin{tabular}[l]{@{}l@{}}without\\ Motion\end{tabular} & \begin{tabular}[l]{@{}l@{}}with\\ Motion\end{tabular} \\ \midrule
LSTM~\cite{varol2018long} & 82.4 & 92.7 \\ 
I3D~\cite{carreira2017quo} & 84.5$^*$ & 93.4$^*$ \\ 
TSN~\cite{wang2016temporal} & 87.3$^*$ & 94.2 \\ 
TLE~\cite{diba2017deep} & 86.9$^*$ & 95.6 \\ 
CoViAR~\cite{wu2018compressed} & 89.7 & 94.9 \\
RF~\cite{piergiovanni2019representation} & 85.5 & 94.5\\
\midrule
ActionVLAD$^{**}$ & 81.81$^*$ & 87.10$^*$\\
Ours$^{**}$ & 89.7$^*$ & 92.2$^*$ \\
Ours & 88.9 & 91.0 \\ \bottomrule
\end{tabular}%
}
\caption{Comparison with the state-of-the-art on UCF101.\newline
$^*$ are results from split1. $^{**}$ uses an image input size of $256\times256$. ActionVLAD uses a 1~FPS sampling rate and 7-second window. The other settings of ActionVLAD is the same as the NetVLAD settings used in Section~\ref{compare_bovw}.}
\label{ucf_soa}
\end{table}
Table~\ref{ucf_soa} lists the state-of-the-art approaches on UCF101 dataset. 
Without motion features our approach outperforms two stream, I3D, LSTM, TSN, and TLE. 
Including motion features, the proposed approach is around 4\% less accurate than the best approach. 
This shows the importance of using good motion features;  we used fewer frames and faster dense optical flow, which might be less accurate. 
We also emphasize the importance of using audio features like VGGish, since they are low-dimensional and fast to extract. 
We also trained ActionVLAD, which uses NetVLAD to encode spatial-temporal information, at a 1~FPS sampling rate, and found that TBEN outperformed ActionVLAD under this limited computational budget.

\section{Conclusion}
\label{sec:conclusion}
\noindent We proposed Temporal Bilinear Encoding Network (TBEN) for encoding long range spatial-temporal information. We compose two constraints in the experiments, working at 1 FPS and using a single GPU. We embedded the label hierarchy in the TBEN and conducted experiments on FGA240. We improved upon the state-of-the-art by applying TBEN on extracted deep visual features and deep audio features and using a hierarchical label loss. 
The result is significantly faster than the state-of-the-art at training and inference time.
We also use TBEN on UCF101 to compute an audio-visual embedding. 
Unfortunately, as there is no hierarchy information in this dataset, we could not use the hierarchical loss. 
Including (1) the mid-frame selection strategy, and (2) optical flow, gave close to state-of-the-art results, with only approx.~3\% less accuracy than the far more computationally expensive models.

\section*{\uppercase{Acknowledgements}}

\noindent This work has emanated from research conducted with the financial support of Science Foundation Ireland (SFI) under grant number SFI/15/SIRG/3283 and SFI/12/RC/2289\_P2.

\bibliographystyle{apalike}
{\small
\bibliography{example}}

\end{document}